\documentclass{article}
\usepackage{spconf,amsmath,epsfig}
\usepackage{spconf,amsmath,graphicx,colortbl}
\usepackage{xcolor}
\usepackage{hyperref}
\hypersetup{hidelinks,
	colorlinks=true,
	allcolors=black,
	pdfstartview=Fit,
	breaklinks=true}
\let\OLDthebibliography\thebibliography
\renewcommand\thebibliography[1]{
  \OLDthebibliography{#1}
  \setlength{\parskip}{0pt}
  \setlength{\itemsep}{0pt plus 0.3ex}
  \setlength{\abovecaptionskip}{0cm}
}

\begin{document}\sloppy

\def\x{{\mathbf x}}
\def\L{{\cal L}}

\title{Latent Degradation Representation Constraint for Single Image Deraining}
%
\name{Yuhong He$^{*1}$, Long Peng$^{*2}$, Lu Wang†$^{1}$, Jun Cheng$^{3}$  
\thanks{
    }
   }


\maketitle

%
%

\setlength{\abovecaptionskip}{0cm}
\ninept
\maketitle
\begin{abstract}
Since rain shows a variety of shapes and directions, learning the degradation representation is extremely challenging for single image deraining. Existing methods mainly propose to designing complicated modules to implicitly learn latent degradation representation from rainy images. However, it is hard to decouple the content-independent degradation representation due to the lack of explicit constraint, resulting in over- or under-enhancement problems. To tackle this issue, we propose a novel Latent Degradation Representation Constraint Network (LDRCNet) that consists of the Direction-Aware Encoder (DAEncoder), Deraining Network, and Multi-Scale Interaction Block (MSIBlock). Specifically, the DAEncoder is proposed to extract latent degradation representation adaptively by first using the deformable convolutions to exploit the direction property of rain streaks. Next, a constraint loss is introduced to explicitly constraint the degradation representation learning during training. Last, we propose an MSIBlock to fuse with the learned degradation representation and decoder features of the deraining network for adaptive information interaction to remove various complicated rainy patterns and reconstruct image details. Experimental results on five synthetic and four real datasets demonstrate that our method achieves state-of-the-art performance. The source code is available at \href{https://github.com/Madeline-hyh/LDRCNet} {https://github.com/Madeline-hyh/LDRCNet}.

\end{abstract}
\vspace{-5pt}
\section{Introduction}
\label{sec:intro}

Images captured in rainy scenes will introduce artifacts like rain streaks and rain accumulation, which would lead to a loss of image detail and contrast. This will degrade the performance of outdoor computer vision systems, such as autonomous driving and video surveillance \cite{driving}. Therefore, restoring rainy images is an essential pre-processing step, and it has drawn much attention in recent years \cite{peng2021ensemble, ye2021closing, yi2021structure, wang2023decoupling,li2023ntire,yan2023textual}. However, single image deraining is still very challenging due to the difficulty in learning the degradation representation of rain streaks under various scenarios \cite{benchmark}.

\begin{figure}[tb]
\small
\centering
\begin{minipage}[b]{1.0\linewidth}
  \centering
\includegraphics[width=1.0\linewidth]
{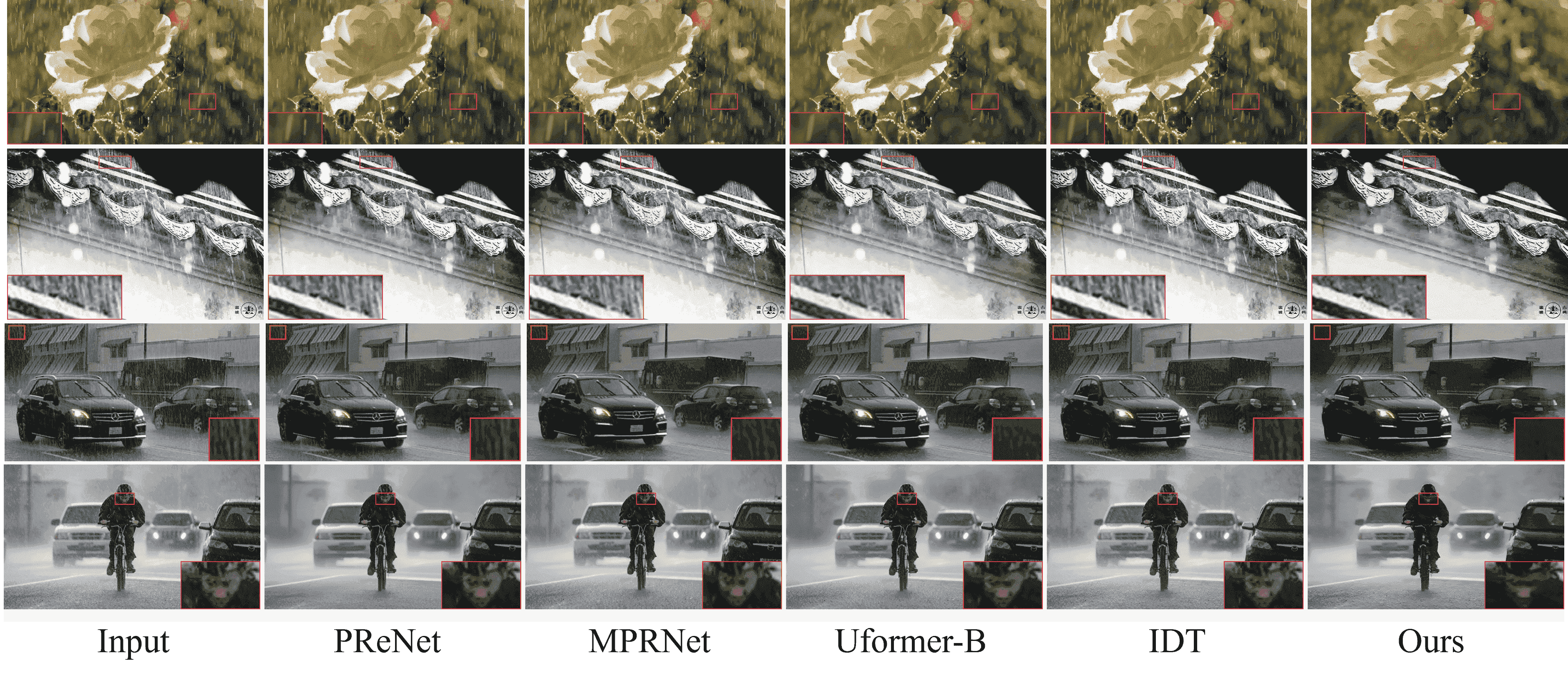}
\end{minipage}
\vspace{-20pt}
\caption{Visual comparison on the real datasets, including Real15 \cite{JORDER} and Real300 \cite{Real300}. Our method reconstructs credible textures with visually pleasing content. }
\label{Real15}
\vspace{-20pt}
\end{figure}
Traditional methods usually solve this problem by calculating a mathematical statistic to obtain diverse priors by exploring the physical properties of rain streaks \cite{Gauss, sparsity}. However, traditional methods have difficulty dealing with complex rainy images in real-world scenarios. Therefore, many deep learning-based methods have recently been proposed for single image deraining \cite{SPANet, peng, he2022deep, uformer, IDLIR,wang2022self}. For example, according to the direction property of rain streaks, Wang \textit{et al.} \cite{SPANet} proposed a spatial attentive network to remove rain streaks in a local-to-global manner. Ma \textit{et al.} \cite{IDLIR} proposed to integrate degradation learning by an iterative framework. Albeit these methods have made significant progress, they still suffer from performance bottlenecks. This is because rain streaks and background are tightly coupled while existing methods focus on designing various modules to learn the degradation representation implicitly and are unable to decouple content-independent degradation representation, which would result in insufficient rain streak residual (\emph{i.e.} under-enhancement) or smooth image textures (\emph{i.e.} over-enhancement). Thus, explicit latent degradation representation learning is critical for single image deraining, which can handle spatially varying rainy patterns in different scenarios and adaptively enhance the structural information to decouple rainy images.

\begin{figure*}[tb]
\small
\centering
\begin{minipage}[b]{1.0\linewidth}
  \centering
\includegraphics[width=0.95\linewidth]{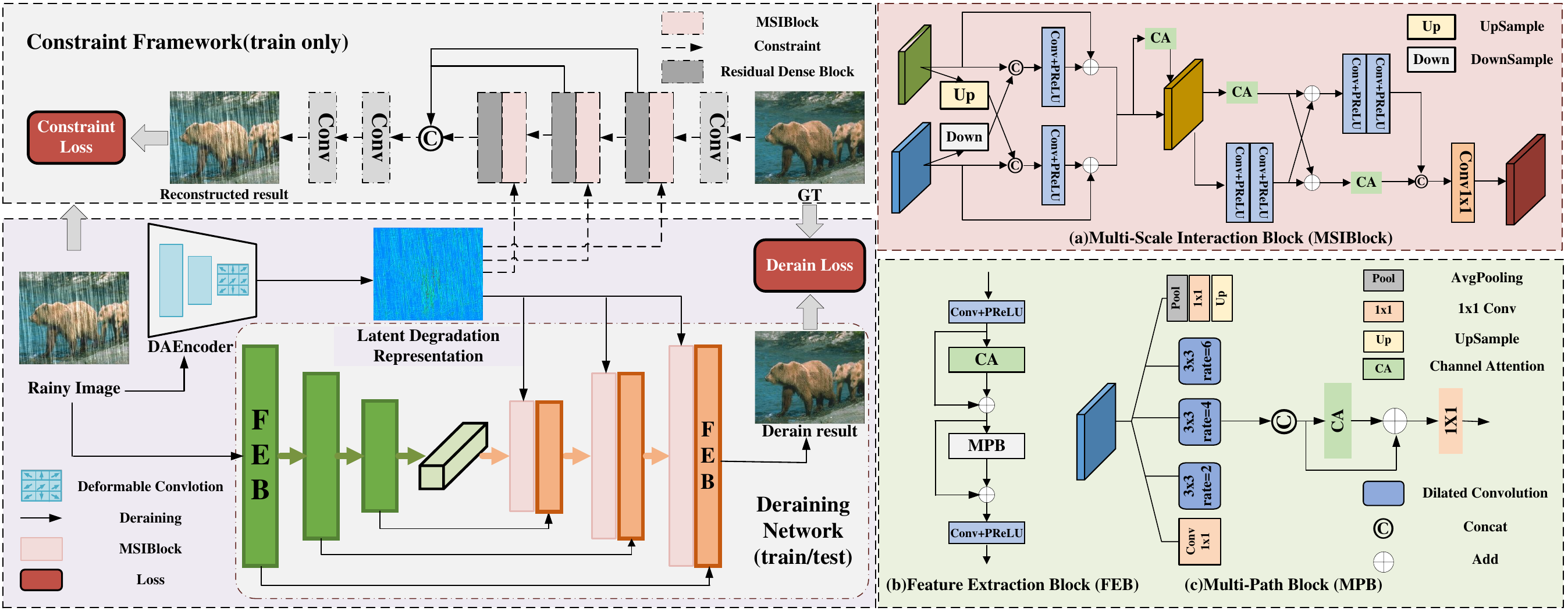}
\end{minipage}
\caption{Our proposed LDRCNet consists of the Direction-Aware Encoder (DAEncoder), Deraining Network, and Multi-Scale Interaction Block (MSIBlock). The constraint framework is proposed to provide explicit supervision to DAEncoder.}
\label{overall}
\vspace{-15pt}
\end{figure*}

To achieve this goal, we propose a novel Latent Degradation Representation Constraint Network (LDRCNet) for explicit degradation learning to remove rain streaks and enhance image details adaptively. Specifically, we propose a direction-aware encoder (DAEncoder) to extract the latent degradation representation by first utilizing deformable convolution \cite{dai2017deformable}, which is based on the directional consistency of rain streaks in the local region \cite{rain_direction} and the directional perception capacity of deformable convolution \cite{deform_Kulkarni_2023_WACV, deform_guan2022memory}. To explicitly supervise the latent degradation representation, we introduce a constraint framework by utilizing the rain-free image and the latent degradation representation learned by the DAEncoder to reconstruct the corresponding rainy image during training. In this way, the latent degradation representation can disentangle the content-independent representation of rain degradation by optimizing the loss between the reconstructed result and the rainy image. To help the deraining network decouple the rain degradation and clean background, we propose a Multi-Scale Interaction Block (MSIBlock) to fuse the content-independent degradation representation and content-dependent decoder features of the deraining network. Such adaptive information interaction enables the deraining network to handle various and complicated rainy patterns effectively and reconstruct the details of images, and our deraining network only adopts a simple yet effective U-Net architecture without fancy design. The main contributions can be summarized as follows:

(1) We propose a novel Latent Degradation Representation Constraint Network (LDRCNet), which utilizes the Direction-Aware Encoder (DAEncoder) to learn the direction-aware degradation representation of rain streaks and is supervised by our proposed novel explicit constraint framework. 

(2) We propose a novel Multi-Scale Feature Interaction Block (MSIBlock) that fuses learned degradation representation and decoder features of deraining networks to handle complex rain patterns and reconstruct image details adaptively.

(3) Experiments on five synthetic and four real-world datasets demonstrate that LDRCNet achieves state-of-the-art performance with explicit latent degradation representation constraints and adaptive information interaction.

\vspace{-10pt}
\section{PROPOSED METHOD}
\label{sec:typestyle}
In this work, we propose a novel Latent Degradation Representation constraint Network (LDRCNet), as shown in Fig. \ref{overall}. We design a Direction-Aware Encoder (DAEncoder) to perceive the directional properties of rain streaks and extract the latent degradation representation, and a constraint framework is proposed to provide explicit supervision. To effectively take advantage of the learned degradation representation for separating the rain layer and background, a Multi-Scale Feature Interaction Block (MSIBlock) is proposed for both the constraint framework and the deraining network.
\vspace{-10pt}
\subsection{Latent Degradation Representation Learning}
\label{ssec:subhead}
{\bf Direction-Aware Encoder.} Inspired by the observation of the direction consistency of rain streaks in local regions, we propose the DAEncoder that consists of several deformable convolutions \cite{dai2017deformable} to learn multi-scale direction-aware degradation representation $deg=\{{deg}_{1}, {deg}_{2}, {deg}_{3}\}$ from the rainy image $R$. Specifically, deformable convolution can adjust the receptive field of the convolution kernel to adapt to the actual geometric variations of the rain streaks by using learnable offsets, which are used to extract the latent degradation representation of rain streaks with different shapes and directions adaptively. For each location ${p}_{0}$ on the output feature map $y$, deformable convolution can be formulated as:
{\setlength\abovedisplayskip{3pt}
\setlength\belowdisplayskip{3pt}
\begin{equation}
\mathbf{y}\left(\mathbf{p}_{0}\right)=\sum_{\mathbf{p}_{l} \in L} \mathbf{w}\left(\mathbf{p}_{l}\right) \cdot \mathbf{x}\left(\mathbf{p}_{0}+\mathbf{p}_{l}+\Delta \mathbf{p}_{l}\right)
\end{equation}}
where ${p}_{l}$ enumerates the offset relative to the kernel center  ${p}_{0}$, which is in $L=\{(-1,-1),\ldots,(1,1)\}$. For example, $L$ has nine choices for a 3x3 convolution kernel. $\mathbf{w}$ represents the weight of different locations. Compared with ordinary convolution, deformable convolution has an augmented offset $\Delta \mathbf{p}_{l}$, which can be learned from the data to adapt to the different directions of rain streaks. Finally, the learning process of DAEncoder can be expressed as:
{\setlength\abovedisplayskip{5pt}
\setlength\belowdisplayskip{5pt}
\begin{equation}
\{{deg}_{1}, {deg}_{2}, {deg}_{3}\} = E(R)
\end{equation}
}
\noindent{\bf Constraint Framework.} 
Most existing methods implicitly learn the degradation representation of rain streaks, which may result in insufficient rain residual or image textures being smoothed. To tackle this issue, we propose a constraint framework. Specifically, the framework learns to reconstruct the corresponding rainy image $\hat{R}$ from the input of the rain-free image $B$ and the multi-scale degradation representation. We introduce a constraint loss $\mathcal{L}_C$ to optimize the loss between the reconstructed result and the original rainy image. In this way, the latent degradation representation learned by the DAEncoder can be supervised by the original rainy image $R$, which tends to be the content-independent representation of rain degradation. The constraint framework structure is based on several Residual Dense Blocks (RDB) \cite{zhang2018residual} and MSIBlock, as shown in Fig . \ref{overall}. The RDB can maximize the information flow and realize feature reuse. The MSIBlock is described in detail in Section 2.2. 

The constraint framework $C$ is only used in the training phase and the learning process and the loss function can be expressed as:
{\setlength\abovedisplayskip{3pt}
\setlength\belowdisplayskip{0pt}
\begin{equation}
\begin{array}{l}
\mathcal{L}_C(R,\hat{R}) = \mathcal{L}(R, C(B, deg))
\end{array}
\end{equation}
}
\vspace{-15pt}
\begin{table*}[tb]
\small
\centering
\caption{Quantitative PSNR($\uparrow$) and SSIM($\uparrow$) comparisons with existing state-of-the-art deraining methods. Average means the average performance of the five benchmark datasets. The \textbf{bold} and \underline{underline} represent the best and second-best performance. }
\label{quantitative} 
\resizebox{1.0\linewidth}{!}{
\begin{tabular}{ccccccccccc|
>{\columncolor[HTML]{EFEFEF}}c 
>{\columncolor[HTML]{EFEFEF}}c }
\hline
                & \multicolumn{2}{c}{Test100} & \multicolumn{2}{c}{Rain100H} & \multicolumn{2}{c}{Rain100L} & \multicolumn{2}{c}{Test1200} & \multicolumn{2}{c|}{Test2800} & \multicolumn{2}{c}{\cellcolor[HTML]{EFEFEF}Average} \\
Metrics         & PSNR         & SSIM         & PSNR          & SSIM         & PSNR          & SSIM         & PSNR          & SSIM         & PSNR          & SSIM          & PSNR                     & SSIM                     \\ \hline
DerainNet \cite{DerainNet}       & 22.77        & 0.810        & 14.92         & 0.592        & 27.03         & 0.884        & 23.38         & 0.835        & 24.31         & 0.861         & 22.48                    & 0.796                    \\
SEMI \cite{SEMI}           & 22.35        & 0.788        & 16.56         & 0.486        & 25.03         & 0.842        & 26.05         & 0.822        & 24.43         & 0.782         & 22.88                    & 0.744                    \\
DIDMDN \cite{DIDMDN}          & 22.56        & 0.818        & 17.35         & 0.524        & 25.23         & 0.741        & 29.95         & 0.901        & 28.13         & 0.867         & 24.64                    & 0.770                    \\
URML \cite{URML}            & 24.41        & 0.829        & 26.01         & 0.832        & 29.18         & 0.923        & 30.55         & 0.910        & 29.97         & 0.905         & 28.02                    & 0.880                    \\
RESCAN \cite{RESCAN}          & 25.00        & 0.835        & 26.36         & 0.786        & 29.80         & 0.881        & 30.51         & 0.882        & 31.29         & 0.904         & 28.59                    & 0.858                    \\
SPANet \cite{SPANet}          & 23.17        & 0.833        & 26.54         & 0.843        & 32.20         & 0.951        & 31.36         & 0.912        & 30.05         & 0.922         & 28.66                    & 0.892                    \\
PReNet \cite{PreNet}          & 24.81        & 0.851        & 26.77         & 0.858        & 32.44         & 0.950        & 31.36         & 0.911        & 31.75         & 0.916         & 29.43                    & 0.897                    \\
MSPFN \cite{MSPFN}           & 27.50        & 0.876        & 28.66         & 0.860        & 32.40         & 0.933        & 32.39         & 0.916        & 32.82         & 0.930         & 30.75                    & 0.903                    \\
PCNet \cite{PCNet}           & 28.94        & 0.886        & 28.38         & 0.870        & 34.19         & 0.953        & 31.82         & 0.907        & 32.81         & 0.931         & 31.23                    & 0.909                    \\
MPRNet \cite{MPRNet}          & \underline{30.27}        & 0.897        & \underline{30.41}         & 0.890        & 36.40         & 0.965        & \textbf{32.91}         & 0.916        & \underline{33.64}         & \underline{0.938}        & \underline{32.73}                   & 0.921                    \\
IDLIR \cite{IDLIR}           & 28.33        & 0.894        & 29.33         & 0.886        & 35.72         & 0.965        & 32.06         & 0.917        & 32.93         & 0.936         & 31.67                    & 0.920                    \\
Uformer-B \cite{uformer}       & 28.71        & 0.896        & 27.54         & 0.871        & 35.91         & 0.964        & 32.34         & 0.913        & 30.88         & 0.928         & 31.08                    & 0.914                    \\
IDT \cite{IDT}             & 29.69        & \underline{0.905}        & 29.95         & \textbf{0.898}        & \textbf{37.01}        & \textbf{0.971}       & 31.38         & 0.908        & 33.38         & 0.937         & 32.28                    & \underline{0.924}                    \\
Semi-SwinDerain \cite{semi-derain} & 28.54        & 0.893        & 28.79         & 0.861        & 34.71         & 0.957        & 30.96         & 0.909        & 32.68         & 0.932         & 31.14                    & 0.910                    \\
DAWN \cite{jiang2023dawn}         & 29.86        & 0.902        & 29.89         & 0.889        & 35.97         & 0.963        & 32.76         & \textbf{0.919}       & -             & -             & 32.12                    & 0.918                    \\
LDRCNet(Ours)   & \textbf{31.17}       & \textbf{0.914}        & \textbf{30.63}         & \underline{0.897}        & \underline{36.83}        & \underline{0.968}       & \underline{32.89}         & \underline{0.917}        & \textbf{33.74}         & \textbf{0.940}         & \textbf{33.05}                   & \textbf{0.927}                    \\ \hline
\end{tabular}
}
\vspace{-5pt}
\end{table*}

\begin{table*}[tb]
\vspace{-10pt}
\caption{Quantitative NIQE($\downarrow$)/BRISQUE($\downarrow$) performance comparisons on the real-world datasets.}
\centering
\label{NIQE} 
\setlength{\tabcolsep}{5pt}
\resizebox{1.0\linewidth}{!}{
\begin{tabular}{ccccccccc}
\hline
Datasets    & UMRL \cite{URML}          & PReNet \cite{PreNet}       & MSPFN \cite{MSPFN}         & PCNet \cite{PCNet}        & MPRNet \cite{MPRNet}        & Uformer-B \cite{uformer}    & IDT \cite{IDT}          & LDRCNet(Ours) \\ \hline
Real15 \cite{JORDER}      & 16.60/24.09 & 16.04/25.29 & 17.03/23.60 & 16.19/25.61 & 16.48/23.92 & 15.71/\textbf{18.67} & 15.60/28.35 & \textbf{14.60}/18.68 \\
Real300 \cite{Real300}     & 15.78/26.39 & 15.12/23.57 & 15.34/28.27 & 15.47/28.99 & 15.08/28.69 & 14.90/24.35 & 15.45/\textbf{23.48} & \textbf{14.67}/28.93 \\
RID \cite{RID}         & 12.19/40.67   & 12.45/38.89    & 11.74/41.88          & 12.03/40.76    & 11.79/44.04   & 11.54/36.38   & 12.49/\textbf{35.99}   & \textbf{11.49}/37.79             \\
RIS \cite{RID}        & 16.30/47.98   & 16.74/48.83    & \textbf{15.73}/47.45 & 16.19/49.00    & 16.88/52.11   & 16.00/\textbf{45.06}  & 17.45/49.86   & 15.76/49.08          \\     
\hline    
\end{tabular}
}
\vspace{-10pt}
\end{table*}
\subsection{Deraining with Learned Degradation Representation}
\label{ssec:subhead}
To let the deraining network adaptively decouple rain streaks in complicated scenes and reconstruct the details of the images, we propose an MSIBlock to interact the learned content-independent degradation representation of rain streaks with the content-dependent decoder features of the deraining network for adaptive feature fusion.

\noindent{\bf Deraining Network.}  
Rich multi-scale representation has fully demonstrated its effectiveness in removing rain streaks \cite{MSPFN}. Therefore, we use a simple yet effective U-Net architecture as the deraining network to extract feature maps at different scales. To retrieve more contextual information, we further propose a Multi-Path Block (MPB) in each feature extraction layer to aggregate more features of rain streaks with a larger receptive field. The MPB has several branches to enlarge the receptive field in parallel, as shown in Fig. \ref{overall} (c). In particular, the MPB first utilizes 1 $\times$ 1 convolution, avgpooling, and dilated convolutions with different dilation rates to capture the multi-scale structure of rain streaks while maintaining negligible parameter increase. Then, all of the feature maps are concatenated, and Channel Attention (CA) is used to adaptively focus on the important feature information. Last, the convolution is used to output the final result $\hat{B}$. The decoder structure is the same as the encoder, and the learned content-independent degradation representation is embedded in the decoder feature by MSIBlock.

After obtaining the pre-trained DAEncoder $E$, we freeze it and retrain the deraining network $D$, which can be expressed as:
{\setlength\abovedisplayskip{3pt}
\setlength\belowdisplayskip{3pt}
\begin{equation}
\begin{array}{l}
\mathcal{L}_D(B,\hat{B}) = \mathcal{L}(B, D(R, deg))
\end{array}
\end{equation}
}
where $R$ and $B$ denote the input of the original rainy image and its corresponding clean image. 

\noindent{\bf Multi-Scale Interaction Block.} To make full use of the multi-scale degradation representation learned by DAEncoder to enhance the structural details and decouple the rain streaks, we propose to embed latent degradation representation into the deraining network. One simple solution is concatenation, but such an operation cannot effectively exploit learned degradation representation to extract complicated rain streaks and may cause optimization interference. Therefore, we propose the MSIBlock for adaptive information interaction. Specifically, we first utilize convolutions to align the content-independent degradation representation $deg$ and content-dependent decoder features $\mathcal{F}_{r}$ of the deraining network, and then Channel Attention (Att) is used to adaptively enhance the important interactive information, as shown in Fig. \ref{overall} (a). Last, diverse combinations of Residual Blocks (RB) can further reconstruct the detail of the image. The MSIBlock can be denoted as:
{\setlength\abovedisplayskip{3pt}
\setlength\belowdisplayskip{3pt}
\begin{equation}
\begin{array}{l}
\mathcal{F}_{c}=\operatorname{Att}\left(\operatorname{Concat}\left(\operatorname{Conv}\left(\mathcal{F}_{r}\right), \operatorname{Conv}\left(deg\right)\right)\right) \\
\widehat{\mathcal{F}}=\operatorname{Concat}\left(\operatorname{RB}\left(\mathcal{F}_{c}\right)\right)
\end{array}
\end{equation}
}
where $\mathcal{F}_{c}$ and $\widehat{\mathcal{F}}$ denotes the concated and the output feature. The MSIBlock is also used in the constraint framework for feature fusion. 

\vspace{-10pt}
\subsection{Loss Function}
\label{ssec:subhead}
The total training loss $\mathcal{L}_{\text {total }}$ can be formulated as follows:
{\setlength\abovedisplayskip{5pt}
\setlength\belowdisplayskip{5pt}
\begin{equation}
\mathcal{L}_{\text {total }}= \lambda_{1}\mathcal{L} _D(B,\hat{B})+\lambda_{2}\mathcal{L}_C(R,\hat{R})
\end{equation}
}
where $\lambda_{1}$ and $\lambda_{2}$ denote the balancing parameters. Following previous work \cite{DIDMDN, MSPFN, VIT, peng}, we use widely-used MSELoss as $\mathcal{L}$.

	
\begin{figure*}[tb]
\setlength{\abovecaptionskip}{-10cm}
\setlength{\belowcaptionskip}{-10cm} 
\centering
\includegraphics[width=0.95\textwidth,height=0.45\textwidth]{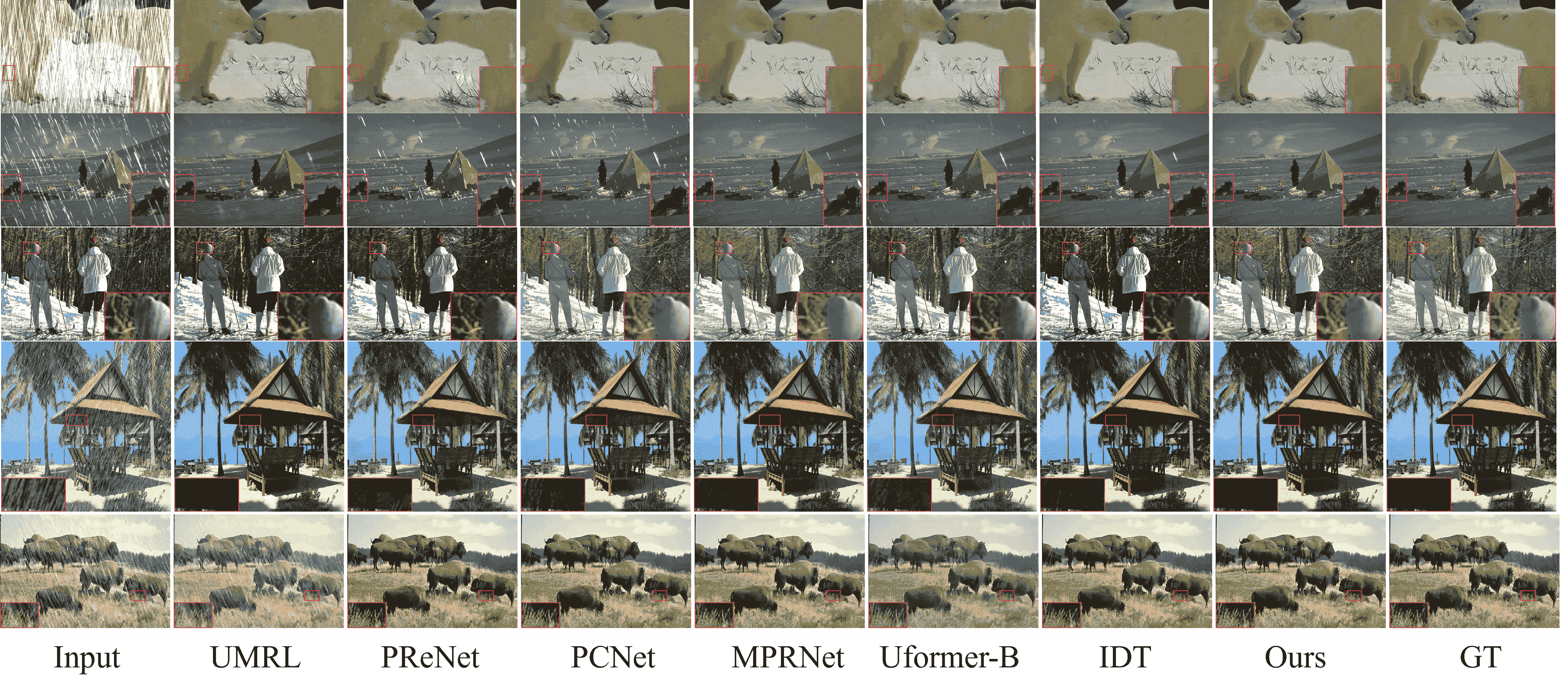}
\centering
\setlength{\abovecaptionskip}{-5pt} 
\vspace{-15pt}
\caption{Visual comparison on the Rain100H \cite{JORDER}, Rain100L \cite{JORDER}, Test100 \cite{Rain800}, Test1200 \cite{DIDMDN}, and Test2800 \cite{DerainNet} datasets. }
\label{visual}
\vspace{-10pt}
\end{figure*}
\vspace{-10pt}

\section{Experiments}
\label{sec:illust}

\subsection{Implementation Details}
\label{ssec:subhead}

In our experiment, we set the training patch size to 256 $\times$ 256 and set $\lambda_{1}$ and $\lambda_{2}$ to 1 and 1. We use the Adam optimizer with an initial learning rate $3 \times 10^{-4}$ for training our methods on four NVIDIA GeForce RTX 3090 GPUs at Pytorch. And the learning rate of Adam is steadily decreased to $1 \times 10^{-6}$  using the cosine annealing strategy. 

\vspace{-5pt}
\subsection{Datasets and Compared Methods}
\label{ssec:subhead}
Following \cite{MSPFN, MPRNet}, we conduct experiments on the Rain13k dataset, which contains 13,712 images with rain streaks of various scales and directions for training, and Test100 \cite{Rain800}, Rain100H \cite{JORDER}, Rain100L \cite{JORDER}, Test1200 \cite{DIDMDN} and Test2800 \cite{DerainNet} are used as test data. Real datasets are also considered to test the generalization, including Real15 \cite{JORDER}, Real300 \cite{Real300}, Rain in Driving (RID), and Rain in Surveillance (RIS) \cite{RID}. RID and RIS have a total of 2495 and 2348 images in real scenes, respectively. We compare our method with the existing fifteen state-of-the-art deraining methods. Continuing along the trajectory of previous works \cite{IDT, MPRNet, PreNet}, we use PSNR and SSIM to evaluate the deraining performance of synthetic images, and utilize NIQE and BRISQUE to evaluate the real dataset. 

\vspace{-10pt}
\subsection{Quantitative and Qualitative Experiment}
\label{ssec:subhead}
{\bf Synthetic Scene.} 
To quantitatively demonstrate the superiority of our method, we compare our method with several existing SOTA methods, and the results are shown in Tab. \ref{quantitative}. Our method achieves the best results on the average performance of five test datasets and we also perform visual comparisons, as shown in Fig. \ref{visual}. Compared with existing methods, we can observe that our method removes rain streaks more completely and restores better texture details of the background, while other approaches retain some obvious rain streaks or lose important details of the background. 

\noindent{\bf Real Scene.} We further conduct experiments on four real-world datasets: Real15 \cite{JORDER}, Real300 \cite{Real300}, RID \cite{RID}, and RIS \cite{RID}. Quantitative results of NIQE and BRISQUE are shown in Tab. \ref{NIQE}. Our LDRCNet achieves the best NIQE performance on Real15, Real300, and RID datasets, which demonstrates that our method achieves good robustness and generalization in real-world scenarios. Visual comparisons on Real15 and Real300 are illustrated in Fig. \ref{Real15}.

\vspace{-10pt}
\subsection{Feature Visualization}
\label{ssec:subhead}
\begin{figure}[h]
\small
\centering
\begin{minipage}[b]{1.0\linewidth}
  \centering
\includegraphics[width=0.95\linewidth]{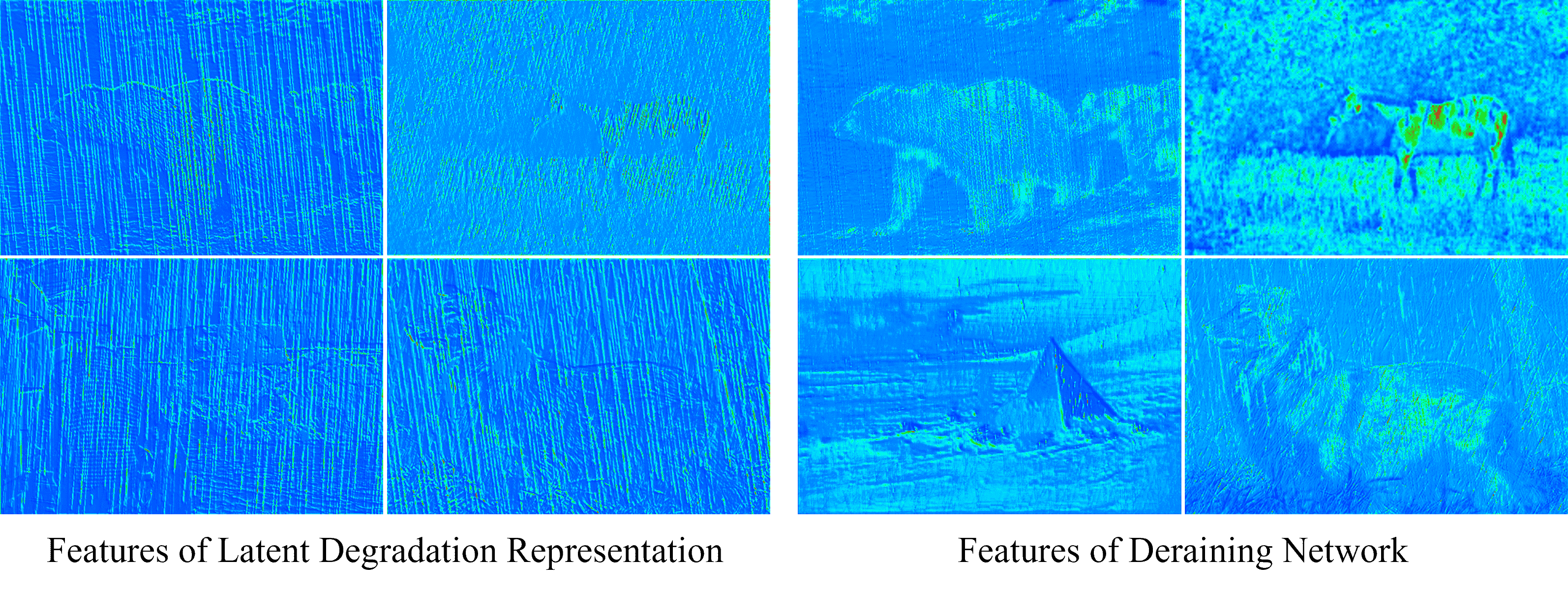}
\end{minipage}
\vspace{-20pt}
\caption{Visualization of the latent feature. }
\label{feature}
\vspace{-20pt}
\end{figure}

To verify the effectiveness of our network structure, we visualize the intermediate feature, as shown in Fig. \ref{feature}. The latent degradation representation with explicit constraints learned content-independent rain degradation, which helps to remove rain streaks. On the contrary, the deraining network learned content-dependent features, which helps to reconstruct the details.

\begin{table}[tb]
\vspace{-13pt}
    \caption{Ablation studies on different settings.}
	\centering
	
	\label{ablation} 
    \setlength{\tabcolsep}{6pt}
    \resizebox{\columnwidth}{!}{
    \begin{tabular}{c|cccccc}
	\hline	
		 & S1 & S2 & S3 & S4 & S5 & Ours\\
    \hline
    PSNR & 27.52 & 28.37 & 29.43 & 29.89 & 29.96 & \textbf{30.60}\\
    SSIM & 0.847 & 0.865 & 0.873 & 0.879 & 0.878 & \textbf{0.897}\\
	\hline
	\end{tabular}
 }
\vspace{-15pt}
\end{table}

\vspace{-5pt}
\subsection{Ablation Studies}
\label{ssec:subhead}
We perform ablation studies on the Rain100H: S1: Without deraining network, and we use convolutions to map the latent degradation representation to rain residual; S2: Without the DAEncoder and the constraint framework; S3: Without constraint framework; S4: Using vanilla convolution to replace the deformable convolution; S5: Using concatenation to replace the MSIBlock. In Tab. \ref{ablation}, we can observe that all components are crucial for our LDRCNet. For example, the performance of the proposed method degrades 1.17dB and 0.019 on PSNR and SSIM without the constraint, demonstrating the superiority of our constraint strategy. The performance degrades 0.71dB and 0.018 on PSNR and SSIM without the deformable convolution, demonstrating that direction-aware information is helpful for decoupling rainy patterns effectively. The performance degrades 0.64dB and 0.019 without the MSIBlock, demonstrating that adaptive information interaction is critical to removing rainy patterns.



\vspace{-5pt}
\section{Conclusion}
\label{conclusion}

In this paper, we propose a novel LDRCNet for explicit degradation learning to remove rain and reconstruct details adaptively. Specifically, we propose a DAEncoder to utilize the directional properties of rain streaks and a constraint network to provide an explicit guide. To make the well-learned latent degradation representation contribute to the deraining network, we propose the MSIBlock for adaptive information interaction, which helps to remove spatially varying rain patterns adaptively. The proposed network is evaluated on five synthetic and four real prevailing datasets, demonstrating it has state-of-the-art performance compared with several representative methods.
\vfill
\pagebreak
\small
\ninept
\bibliographystyle{IEEEbib}
\bibliography{icme2023template}

\end{document}